\documentclass{llncs}
\usepackage{llncsdoc}

\usepackage[utf8]{inputenc} 
\usepackage[T1]{fontenc}    
\usepackage{hyperref}       
\usepackage{url}            
\usepackage{booktabs}       
\usepackage{amsfonts}       
\usepackage{nicefrac}       
\usepackage{microtype}      

\usepackage{amsmath} 
\usepackage{amsfonts}
\usepackage{amssymb} 
\usepackage{mathtools}
\usepackage{graphicx}
\usepackage{subcaption}
\usepackage[linesnumbered,ruled,vlined]{algorithm2e}
\usepackage{ntheorem}
\theoremstyle{empty}
\newtheorem{named}{}
\title{Learning Syllogism with Euler Neural-Networks}

\authorrunning{Dong et al}
\author{Tiansi Dong$^{1}$\thanks{the first and the second authors have the equal contribution} , Chengjiang Li$^{2*}$, Christian Bauckhage$^{1}$\\ Juanzi Li$^{2}$, Stefan Wrobel$^{1}$, Armin B. Cremers$^1$}\institute{
$^1$University of Bonn, Bonn, Germany\\
$^2$Tsinghua University, Beijing, China 
}

\begin{document}

\maketitle

\begin{abstract} 

Traditional neural networks represent everything as a vector, and are able to approximate a subset of logical reasoning to a certain degree. As basic logic relations are better represented by topological relations between regions, we propose a novel neural network that represents everything as a ball and is able to learn topological configuration as an Euler diagram. So comes the name Euler Neural-Network (ENN). The central vector of a ball is a vector that can inherit representation power of traditional neural network. ENN distinguishes four spatial statuses between balls, namely, being disconnected, being partially overlapped, being part of, being inverse part of. Within each status, ideal values are defined for efficient reasoning. A novel back-propagation algorithm with six Rectified Spatial Units (ReSU) can optimize an Euler diagram representing logical premises, from which logical conclusion can be deduced. In contrast to traditional neural network, ENN can precisely represent all 24 different structures of Syllogism. Two large datasets are created: one extracted from WordNet-3.0 covers all types of Syllogism reasoning, the other extracted all family relations from DBpedia. Experiment results approve the superior power of ENN in logical representation and reasoning. Datasets and source code are available upon request.
\end{abstract}

\section{Introduction}


Deep Learning \cite{dl2016} has solved a variety of difficult AI tasks, e.g., gaming \cite{alphaGo2017}, machine translation, object recognition, robotics \cite{LeCunNature15}. Vectors are used by deep neural-networks to represent  words, sentences, texts, images, videos, and are able to simulate a number of functions of the associative memory (System 1 of mind) \cite{Kahneman11}, and approximate logical reasoning (System 2 of mind) \cite{Bengio2019}. On the other hand, regions are taken as primitive for  commonsense spatial reasoning  \cite{Clarke81,Laguna22a,DongJPL,Randell92a,whitehead29}, also used for logical reasoning \cite{Stapleton10,Venn1880} and cognitive modeling \cite{Smith94}. Using regions as inputs of neural-networks can date back to \cite{minsky88} in terms of {\em diameter-limited perceptrons}, and received continued interests (to increase the power of reasoning) in terms of  {\em  Poincar\'e ball} \cite{Nickel17}, {\em sphere} \cite{JuanziEmnlp18}, {\em N-ball} \cite{dong19iclr}, {\em hyperbolic disks} \cite{hyperbolicDAG19}, {\em boxes} \cite{ren2020}, or using vector plus a bounded distance \cite{mzadeh15}. However, current reasoning still can not allow logical forms to contain negation, and fails to reason with different structures of syllogism. Here, we propose a novel neural-network architecture, namely, Euler Neural-Network (ENN) that takes high dimensional ball as inputs and is able to learn topological configurations of balls as Euler diagram for reasoning. 


Advantages of ENN are as follows: (1) it uses central vectors of balls to inherit latent features from traditional neural-networks; (2) it uses topological relations among balls to encode structures among balls; (3) it uses a map of spatial transition as an innate structure within the network; (4) objective functions are dynamically optimized by the neighborhood transition from the input relation to the target relation; (5) ideal values within topological relations are parameterized not only to realize efficient reasoning but also to optimize visualization. Two large datasets are created for the reasoning of syllogism, and the reasoning of family relations as an example of reasoning with part-whole relations  \cite{Hinton1990}. In contrast to existing works, ENN can precisely represent  all 24 styles of syllogism, and all family relations. Our experiments show that ENN reaches 100\% accuracy in reasoning with syllogism only having three statements. In reasoning with family relation without gender information, the accuracy slightly decreases along with the number of statements. By utilizing pre-trained latent feature vectors, ENN is able to reasoning with family relations with gender information.  

The rest of the paper is structured as follows: Section 2 surveys a number of related work. Section 3 proposes Euler Neural Network, including its architecture, dynamic loss functions, and relations to traditional neural networks. Section 4 presents our experiments in syllogism, and reasoning with family relations. Section 5 concludes the paper and lists some on-going researches.

\section{Euler neural network}


We propose a simple extension of classic neural-networks which promotes vectors into balls and uses topological transition map as its inner structure for spatial optimization. This enables the novel neural-network to learn ball configurations as Euler diagram for logical reasoning. So comes the name Euler Neural-Network (ENN), as illustrated in Figure~\ref{fig:stran}.  In ENN, an entity $w$ is represented as an $n+1$ dimensional vector $\vec{w}$ and is interpreted as an $n$ dimensional ball with the central vector $\vec{O}_w=[w_1, \dots, w_n]$, 
 and the length of the radius $r_w = e^{w_{n+1}} >0$.  We defined ball $(\vec{O}_w, r_w)$ as an open space. That is, a point $p$ is inside ball $(\vec{O}_w, r_w)$, if and only if $\|p-\vec{O}_w\| < r_w$. ENN optimizes the relation between ball $\vec{v}$ and ball $\vec{w}$ to the target relation $\mathbf{T}$. The default value of $\mathbf{T}$ can be a random choice between $\mathbf{P}$ and $\mathbf{\bar{P}}$, so that ENN will optimize the relation between two input balls to either $\mathbf{P}$ to $\mathbf{\bar{P}}$. This will result in the equal relation: $\mathbf{P}(\vec{w},\vec{v})\wedge\bar{\mathbf{P}}(\vec{w},\vec{v})\rightarrow\mathbf{E}(\vec{w},\vec{v})$ and can be measured by the $\cos$ similarity between their central vectors: $\mathbf{E}(\vec{w},\vec{v})\rightarrow\cos(\vec{O}_w, \vec{O}_v) = 1$. That is, ENN is degraded into a traditional neural-network.

\subsection{Spatial predicates}



\begin{figure} 
\centering
\includegraphics[width=0.8\textwidth]{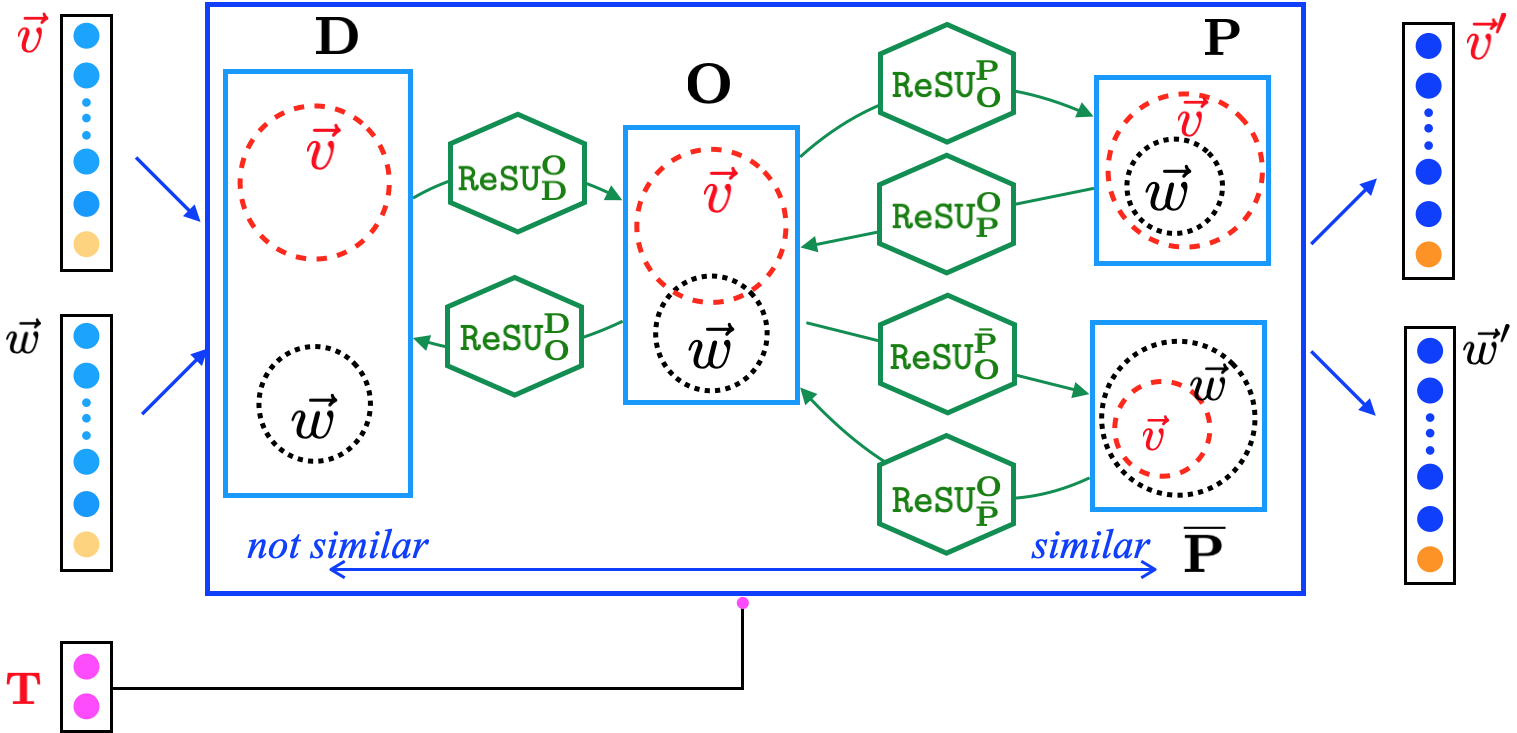} 
\caption{Euler Neural-Network having four spatial statuses, three neighborhood relations, and six Rectified Spatial Units (ReSU)}
\label{fig:stran}
\end{figure}

Given two balls $\vec{w}$ and $\vec{v}$, we define $\mathbf{D}(\vec{w}, \vec{v})$ as a spatial predicate that returns true, if and only if  $\vec{w}$ disconnects from $\vec{v}$. This can be measured by subtracting the sum of their radii from the distance between their central vectors.  
\[
  \mathbf{D}(\vec{w},\vec{v})= 
    true, \mbox{ if and only if }  \|\vec{O}_w - \vec{O}_v\| - (r_w + r_v) \ge 0    
\]

We define $\mathbf{O}(\vec{w}, \vec{v})$ as a spatial predicate that returns true, if and only if  $\vec{w}$ is partially overlapped with $\vec{v}$. This can be determined by checking whether the distance between their central vectors is greater than the difference between their radii, meanwhile less than the sum of their radii.  
\[
  \mathbf{O}(\vec{w},\vec{v})= 
    true, \mbox{ if and only if }    |r_w - r_v| <\|\vec{O}_w - \vec{O}_v\| < r_w + r_v
\]

Ball $\vec{w}$ is part of ball $\vec{v}$, $\mathbf{P}(\vec{w}, \vec{v})$ or  $\mathbf{\bar{P}}(\vec{v}, \vec{w})$, if the distance between their central vectors plus the radius of $\vec{w}$ is less than or equals to the radius of  $\vec{v}$. The co-inside relation (or, the equal relation ($\mathbf{E}$) \cite{whitehead29,Clarke85,Randell92a,DongJPL}) is included by both the $\mathbf{P}$ relation and the $\bar{\mathbf{P}}$ relation.  
\[
  \mathbf{P}(\vec{w},\vec{v})= \mathbf{\bar{P}}(\vec{v},\vec{w}) = 
    true, \mbox{ if and only if }     \|\vec{O}_w - \vec{O}_v\| + r_w \le  r_v
\]

The four spatial predicates are  jointly exhaustive (it holds that $\forall\vec{x}\vec{y}[\mathbf{D}(\vec{x},\vec{y})\vee\mathbf{O}(\vec{x},\vec{y})\vee\mathbf{P}(\vec{x},\vec{y})\vee\bar{\mathbf{P}}(\vec{x},\vec{y})]$)  and pairwise disjoint with one exception that $\forall\vec{x}\vec{y}[\mathbf{P}(\vec{x},\vec{y})\wedge\bar{\mathbf{P}}(\vec{x},\vec{y})\rightarrow\mathbf{E}(\vec{x},\vec{y})]$. Each spatial predicate asserts a spatial status between two input balls. Transitions among neighborhood spatial statuses have been discussed in qualitative spatial reasoning, i.e.,  \cite{Randell92a,Freksa91,DongJPL}. We adopt a lightweight topological transition map of open regions that only consists of three neighborhood relations: $\mathbf{D}\leftrightarrow\mathbf{O}$, $\mathbf{O}\leftrightarrow\mathbf{P}$, and $\mathbf{O}\leftrightarrow\mathbf{\bar{P}}$, as illustrated in Figure~\ref{fig:stran}.


\subsection{Rectified spatial unit (ReSU)}

Rectified activation units have shown better performance than sigmoid or hyperbolic tangent units \cite{BengioReLU,HintonReLU,Ng13}.  Six Rectified Spatial Units (ReSU) are designed to regulate transformations between neighborhood spatial statuses. The ReSU for the transition from $\mathbf{D}$ to $\mathbf{O}$ is defined as \[\mathtt{ReSU}_\mathbf{D}^\mathbf{O}(\vec{w}, \vec{v})\triangleq\max\{0, \|\vec{O}_w - \vec{O}_v\| - (r_w + r_v)\}\]

$\mathtt{ReSU}_\mathbf{D}^\mathbf{O}(\vec{w}, \vec{v})$ is greater than zero, if $\mathbf{D}(\vec{w},\vec{v})$. Decreasing the value of $\mathtt{ReSU}_\mathbf{D}^\mathbf{O}(\vec{w}, \vec{v})$ will push the relation  between $\vec{w}$ and $\vec{v}$ to the relation of {\em being overlapped} ($\mathbf{O}$). That is the `$\mathbf{O}$' in $\mathtt{ReSU}_\mathbf{D}^\mathbf{O}(\vec{w}, \vec{v})$.  
From the relation of being partially overlapped, the relations between two balls can be transformed into either {\em being disconnected} or {\em being part of} (including the inverse relation). We define three Rectified Spatial Units $\mathtt{ReSU}_\mathbf{O}^\mathbf{D}$, $\mathtt{ReSU}_\mathbf{O}^\mathbf{P}$, and $\mathtt{ReSU}_\mathbf{O}^\mathbf{\bar{P}}$ as follows. 
\[
  \mathtt{ReSU}_\mathbf{O}^\mathbf{D}(\vec{w}, \vec{v})\triangleq\max\{0,
    r_w + r_v - \|\vec{O}_w - \vec{O}_v\|\} \]
$\mathtt{ReSU}_\mathbf{O}^\mathbf{D}(\vec{w}, \vec{v})$ is greater than zero, if $\mathbf{O}(\vec{w},\vec{v})$. Decreasing the value of $\mathtt{ReSU}_\mathbf{O}^\mathbf{D}(\vec{w}, \vec{v})$ will push the relation  between $\vec{w}$ and $\vec{v}$ to the relation of {\em being disconnected} ($\mathbf{D}$) between $\vec{w}$ and $\vec{v}$. 
\[
  \mathtt{ReSU}_\mathbf{O}^\mathbf{P}(\vec{w}, \vec{v})\triangleq \max\{0,
     \|\vec{O}_w - \vec{O}_v\| + r_w  - r_v\}\]
$\mathtt{ReSU}_\mathbf{O}^\mathbf{P}(\vec{w}, \vec{v})$ is greater than zero, if $\mathbf{O}(\vec{w},\vec{v})$. Decreasing the value of $\mathtt{ReSU}_\mathbf{O}^\mathbf{P}(\vec{w}, \vec{v})$ will push the relation  between $\vec{w}$ and $\vec{v}$ to the relation that $\vec{w}$ {\em being part of} ($\mathbf{P}$) $\vec{v}$. 
\[
  \mathtt{ReSU}_\mathbf{O}^\mathbf{\bar{P}}(\vec{w}, \vec{v})\triangleq\max\{0,       \|\vec{O}_w - \vec{O}_v\| + r_v -r_w \}
\]
$\mathtt{ReSU}_\mathbf{O}^\mathbf{\bar{P}}(\vec{w}, \vec{v})$ is greater than zero, if $\mathbf{O}(\vec{w},\vec{v})$. Decreasing the value of $\mathtt{ReSU}_\mathbf{O}^\mathbf{\bar{P}}(\vec{w}, \vec{v})$ will  push the relation  between $\vec{w}$ and $\vec{v}$ to the relation that $\vec{v}$ {\em being part of} ($\mathbf{P}$) $\vec{w}$.

The relation of being part of can be transformed into the relation of being partially overlap. We define $\mathtt{ReSU}_\mathbf{P}^\mathbf{O}(\vec{w},\vec{v})$ as follows.
\[
 \mathtt{ReSU}_\mathbf{P}^\mathbf{O}(\vec{w}, \vec{v})\triangleq\max\{0,   
    r_v - r_w - \|\vec{O}_w - \vec{O}_v\|\}\]
We define $\mathtt{ReSU}_\mathbf{\bar{P}}^\mathbf{O}(\vec{w},\vec{v})\triangleq\mathtt{ReSU}_\mathbf{P}^\mathbf{O}(\vec{v},\vec{w}) $

\subsection{Ideal spatial values}

\begin{figure}
\begin{subfigure}[c]{0.5\textwidth}
\includegraphics[width=1\textwidth]{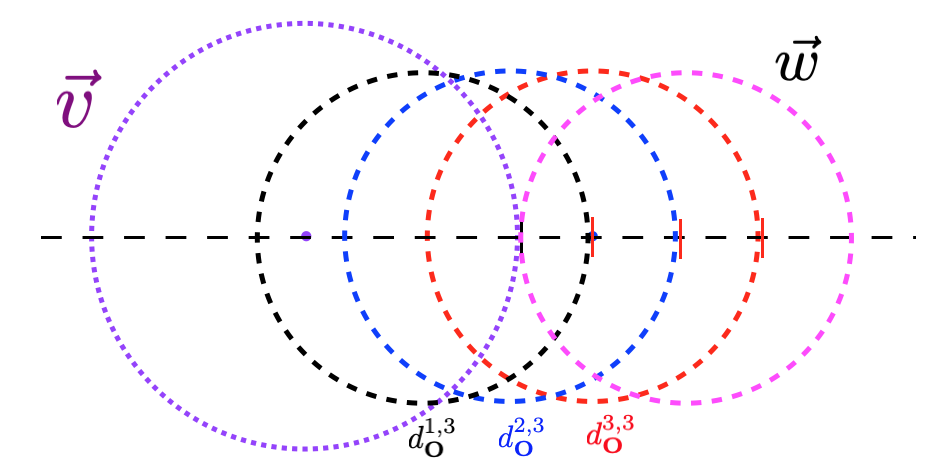}
\subcaption{Three reference relations of being partial overlapped}
\end{subfigure}
\begin{subfigure}[c]{0.5\textwidth}
\centering
\includegraphics[width=0.72\textwidth]{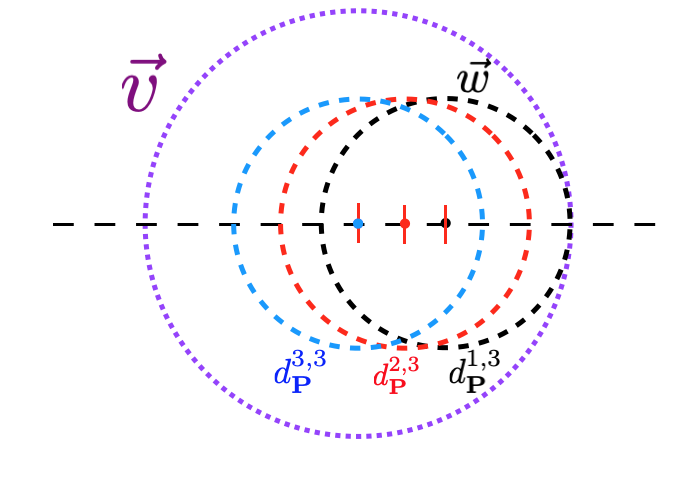}
\subcaption{Three reference relations of being part of}
\end{subfigure}
\caption{ideal values in the spatial categories of {\em being partially overlapped}(a) and {\em being part of}(b)}
\label{fig:ivref}
\end{figure}

In normal back-propagation process, optimization process to transform from $\mathbf{O}(\vec{w}, \vec{w})$ relation to $\mathbf{D}(\vec{w}, \vec{w})$ relation will be stopped, when $\mathtt{ReSU}_\mathbf{O}^\mathbf{D}(\vec{w}, \vec{w})=0$. This makes the disconnected relation between $\vec{w}$ and $\vec{w}$ indistinguishable from the partial overlapping relation between them. This kind of being almost overlapped relation is neither ideal for reasoning nor for visualization. In natural categories, such as color, line orientations, and numbers, people select a subset of members as ``ideal types''\cite{Wertheimer38} or ``cognitive reference points''\cite{Rosch75a}, such as multiples of 10 as ideal numbers, 
vertical, horizontal, and diagonal lines as ideal orientations. 
We define $N$ ideal distance values for the being disconnected relation as follows. 
\[
    d^{k,N}_{\mathbf{D}}(\vec{w}, \vec{v})=      (k-1)(r_w + r_v), \mbox{if}\   \mathbf{D}(\vec{w}, \vec{v})\]
in which $k=1, \dots, N$. We define the spatial function $f^k_{\mathbf{D}}(\vec{w},\vec{v})=(\|\vec{O}_w-\vec{O}_v\|-d^k_{\mathbf{D}}(\vec{w}, \vec{v}))^2$ as the loss function for the training. Fix the radii of $\vec{v}$ and $\vec{w}$, and let $r_v \ge r_w$. We define $N$ ideal distances between the central points of two partially overlapped balls as follows. 
\[
    d^{k,N}_\mathbf{O}(\vec{w}, \vec{v}) =      \frac{2kr_w}{N+1}  + r_v - r_w, \mbox{if}\ \mathbf{O}(\vec{w}, \vec{v})
\]
in which $k=1, \dots, N$. Figure~\ref{fig:ivref}(a) illustrates three ideal partial overlapping relations. $\mathbf{O}$ 
is the transition status between $\mathbf{D}$ and $\mathbf{P}$ ($\mathbf{\bar{P}}$). 
In one extreme case, let $k=0$, $d^{0,N}_\mathbf{O}(\vec{w}, \vec{v})= r_v - r_w$ that means that ball $\vec{w}$ is {\em tangential part of} ball $\vec{v}$; In another extreme case, let $k=N+1$, $d^{N+1,N}_\mathbf{O}(\vec{w}, \vec{v})= r_v + r_w$ that means that ball $\vec{w}$ is {\em exactly disconnected from} ball $\vec{v}$. We define the spatial function $f^{k,N}_{\mathbf{O}}(\vec{w},\vec{v})=(\|\vec{O}_w-\vec{O}_v\|-d^{k,N}_{\mathbf{O}}(\vec{w}, \vec{v}))^2$ as the loss function for the training.

Fix the radii of $\vec{v}$ and $\vec{w}$, and let $r_v > r_w$. We define $N$ ideal distances between the central points of two balls with the condition that one ball is part of the other as follows.
\[
    d^{k,N}_{\mathbf{P}}(\vec{w}, \vec{v}) =      r_v - r_w - \frac{k(r_v-r_w)}{N},\  \mbox{if}\ \ \mathbf{P}(\vec{w}, \vec{v})
\]
in which $k=1, \dots, N$. Figure~\ref{fig:ivref}(b) illustrates three reference part of relations. If $k=1$, $d^{0,N}_{\mathbf{P}}(\vec{w}, \vec{v}) =  r_v - r_w$ that means ball $\vec{w}$ is tangential part of ball $\vec{v}$; If $k=N$, $d^{0,N}_{\mathbf{P}}(\vec{w}, \vec{v}) = 0$ that means two balls are concentric. We define the spatial function $f^{k,N}_{\mathbf{P}}(\vec{w},\vec{v})=(\|\vec{O}_w-\vec{O}_v\|-d^{k,N}_{\mathbf{P}}(\vec{w}, \vec{v}))^2$ as the loss function for the training.
 
Ideal values are invariant, if ball $\vec{w}$ rotates around the central point of ball $\vec{v}$. We define ideal rotation $\chi^{k,N}_{p,q,\vec{v}}(\vec{O}_w)$ as ball $\vec{w}$ rotates $\frac{2k\pi}{N}$ (Euler angle) in the space spanned by the $p^{th}$ and the $q^{th}$ axes around the central point of ball $\vec{v}$. 
\[
  \chi^{k,N}_{p,q,\vec{v}}(\vec{O}_w)= 
\begin{cases}
     (\vec{O}_w[p]-\vec{O}_v[p])\cos(\frac{2k\pi}{N}) - (\vec{O}_w[q]-\vec{O}_v[q])\sin(\frac{2k\pi}{N}) + \vec{O}_v[s], & s=p\\
    (\vec{O}_w[p]-\vec{O}_v[p])\sin(\frac{2k\pi}{N}) + (\vec{O}_w[q]-\vec{O}_v[q])\cos(\frac{2k\pi}{N}) + \vec{O}_v[s],      &  s=q\\
    \vec{O}_w[s], & s\notin \{p, q\}
\end{cases}
\]

\subsection{Learning Euler diagram}

The input of an ENN consists of a sequence $N$-dimensional balls $[\vec{w}_1,\dots,\vec{w}_m]$ and a table of target topological relations $\mathtt{TB}_0$, parameters for ideal values $N_\mathbf{D}, N_\mathbf{O}, N_\mathbf{P}$, the total number of the ideal rotations $M$, and the maximum number of iterations. The output of ENN is the sequence of balls with updated locations and sizes, so that the topological relations among them satisfy the relations defined in $\mathtt{TB}_0$ as much as possible. 
The global optimization procedure is illustrated in Algorithm~\ref{algo:uenn0}.
 
\begin{algorithm}
\DontPrintSemicolon
\KwIn{A sequence of $N$-dimensional balls $[\vec{w}_1, \dots, \vec{w}_m]$}
\KwIn{A table $\mathtt{TB}_0$ of target relations between the balls, $\mathtt{TB}_0(\vec{w}_i,\vec{w}_j)\in\{\mathbf{D}, \mathbf{O}, \mathbf{P}, \mathbf{\bar{P}, \emptyset}\}$ }
\KwIn{$N, N_\mathbf{D},N_\mathbf{P},N_\mathbf{O}, M$, $maxInter$}
\KwOut{Euler diagram for $[\vec{w}_1, \dots, \vec{w}_m]$ satisfying relations in $\mathtt{TB}_0$ as much as possible}  
$Round\gets 0$;$gLoss\gets 1$;\\ 
$V\gets \mbox{initialize, and sort } [\vec{w}_1, \dots, \vec{w}_m]\mbox{ in the decreased order of degrees of } \vec{w}_is$\\
\While{$Round <maxInter\wedge gLoss > 0$}{
    \For{$k =1,\dots, n-1$}{
        \For{$j =k+1,\dots, n$}{
            \If{$\mathtt{TB}_0(V[j],V[k])\neq\emptyset$}
            {
            		$\mathbf{routes} \gets$$ \mbox{\small{get the transition route to}}   \mathtt{TB}_0(V[j],V[k])$\mbox{\small{ in the transition map in Figure~\ref{fig:stran};}}\\
				\For{$(\mathbf{R}_0$, $\mathbf{R}_1) \in \mathbf{routes}$}{
					$z\gets 0$;\\
					\Repeat{$(\mathtt{ReSU}_{\mathbf{R}_0}^{\mathbf{R}_1}(V[j],V[k])=0)\lor (z\ge maxInter)$}{
      						\mbox{\small{update $V[j]$ using back-propagation to reduce the value of $\mathtt{ReSU}_{\mathbf{R}_0}^{\mathbf{R}_1}(V[j],V[k])$;}}\\
 							 \uIf{$\mathtt{ReSU}_{\mathbf{R}_0}^{\mathbf{R}_1}(V[j],V[k])=0$}{
      							\mbox{\small{$l\gets randint(1,N_{\mathbf{R}_1})$}};\\ \mbox{\small{update $V[j]$, till the relation between $V[j]$ and $V[k]$ is $d^{l,N_{\mathbf{R}_1}}_{\mathbf{R}_1}$;}} 
    								}
    						\Else{
    							\small{$gLoss\gets \mbox{compute current global loss with respect to } \mathtt{TB}_0$;}\\
								\For{$p=0,\dots ,N-2$}{
									\For{$q=p+1,\dots ,N-1$}{
										\For{$l=1,\dots,M$}{
											$\vec{O}_{V[j]}\gets \chi^{l,M}_{p,q,V[k]}(\vec{O}_{V[j]})$;\\
									\mbox{\small{$gLoss_1\gets$ compute current global loss with respect to $\mathtt{TB}_0$};}\\
											\If{$gLoss_1<gLoss$}{ $minV\gets V[j]$; $gLoss\gets gLoss_1$}
									}
								  }
								}
    							$V[j]\gets minV$;
    							$z = z + 1$;
    						}
    				}
    			}
             }
    		}
    	}    
    	\If{$gLoss > 0$}{
    	\For{$i=1,\dots, n$}{ 
            \mbox{\small{normal back-propagation operation on $V[i]$ to reduce $gLoss$, till $gLoss$ does not decrease.}}
        }
    }
    $Round\gets Round+1$;}
\Return{$V$}
\caption{{\sc global optimization of Topological Transition}}
\label{algo:uenn0}
\end{algorithm}

Algorithm~\ref{algo:uenn0} randomly initializes balls, and sorts them according to the number of degrees in the decreasing order. The two outer $\mathbf{for}$ loops traverses all target relations in $\mathtt{TB}_0$. To optimize the relation between two balls to the target relation, ENN firstly finds the route to the target relation according to the transition map in Figure~\ref{fig:stran} (the length of a route is either 1 or 2). The optimization of the a route segment is a $\mathbf{repeat}$ loop that starts with a normal back-propagation process \cite{Rumelhart19881}. If ends with $0$, the target relation will further optimized into an ideal value ($d^{l,N_{\mathbf{R}_1}}_{\mathbf{R}_1}$), otherwise current focused ball will rotate with an ideal value ($\chi^{l,M}_{p,q,V[k]}(\vec{O}_{V[j]})$), with which the global loss is the smallest among all possible rotated locations. From that rotate location, the $\mathbf{repeat}$ loop continues the back-propagation. After having traversed $\mathtt{TB}_0$, ENN computes the current global loss $gLoss$. If it is greater than 0, a normal back-propagation will be applied for every ball. After that, ENN continues the two outer $\mathbf{for}$ loops to optimize relations in $\mathtt{TB}_0$ till either $gLoss$ reaches 0 or the maximum iteration number is reached. 


\subsection{Representing all 24 structures of syllogism}

Statements of syllogism consist of four types: (1) all $x$ are $y$; (2) some $x$ are $y$; (3) no $x$ are $y$; (4) some $x$ are not $y$. Type (1) can be interpreted as {\em ball $\vec{x}$ is part of  ball $\vec{y}$} ($\mathbf{P}(\vec{x},\vec{y})$); Type (2) can be interpreted as {\em there is a ball $x_0$ inside of ball $\vec{x}$ such that ball $x_0$ is part of ball $\vec{y}$} ($\exists \vec{x}_0[\mathbf{P}(\vec{x}_0,\vec{x})\wedge\mathbf{P}(\vec{x}_0,\vec{y})]$). This is equivalent to $\mathbf{P}(\vec{x},\vec{y})\vee\mathbf{O}(\vec{x},\vec{y})\vee\bar{\mathbf{P}}(\vec{x},\vec{y})$ and also to $\neg\mathbf{D}(\vec{x},\vec{y})$;  Type (3) can be interpreted as {\em ball $\vec{x}$ disconnects from  ball $\vec{y}$} ($\mathbf{D}(\vec{x},\vec{y})$); Type (4) can be interpreted as {\em there is a ball $x_0$ inside of ball $\vec{x}$ such that ball $x_0$ disconnects from  ball $\vec{y}$} ($\exists \vec{x}_0[\mathbf{P}(\vec{x}_0,\vec{x})\wedge\mathbf{D}(\vec{x}_0,\vec{y})]$). This is equivalent to $\mathbf{D}(\vec{x},\vec{y})\vee\mathbf{O}(\vec{x},\vec{y})\vee\bar{\mathbf{P}}(\vec{x},\vec{y})$ and also to $\neg\mathbf{P}(\vec{x},\vec{y})$. Table~\ref{syllogismList} lists the representations of all 24 different structures of syllogisms that can be precisely represented by ENN.

\begin{table}
  \caption{List of all syllogisms}
  \label{syllogismList}
  \centering
  \hspace*{-1.2em}
  \scalebox{0.9}{
  \setlength{\tabcolsep}{4.0pt} 
  \begin{tabular}{clllr}
    \toprule
   Num & Name     & Premise &Conclusion    & Spatial proposition for Euler diagrams \\
 
    \midrule
   1& Barbara & all $s$ are $m$, all $m$ are $p$ & all $s$ are $p$  & $\mathbf{P}(\vec{s},\vec{m})\wedge\mathbf{P}(\vec{m},\vec{p})\rightarrow\mathbf{P}(\vec{s},\vec{p})$     \\
   2& Barbari & all $s$ are $m$, all $m$ are $p$ & some $s$ are $p$  & $\mathbf{P}(\vec{s},\vec{m})\wedge\mathbf{P}(\vec{m},\vec{p})\rightarrow\neg\mathbf{D}(\vec{s},\vec{p})$      \\
   3&Celarent& all $s$ are $m$, no $m$ is $p$ & no $s$ is $p$& $\mathbf{P}(\vec{s},\vec{m})\wedge\mathbf{D}(\vec{m},\vec{p})\rightarrow\mathbf{D}(\vec{s},\vec{p})$     \\
    4& Cesare& all $s$ are $m$, no $p$ is $m$ & no $s$ is $p$& $\mathbf{P}(\vec{s},\vec{m})\wedge\mathbf{D}(\vec{p},\vec{m})\rightarrow\mathbf{D}(\vec{s},\vec{p})$     \\
   5& Calemes & no $m$ is $s$, all $p$ are $m$ & no $s$ is $p$& $\mathbf{D}(\vec{m},\vec{s})\wedge\mathbf{P}(\vec{p},\vec{m})\rightarrow\mathbf{D}(\vec{s},\vec{p})$ \\
   6& Camestres & no $s$ is $m$, all $p$ are $m$ & no $s$ is $p$& $\mathbf{D}(\vec{s},\vec{m})\wedge\mathbf{P}(\vec{p},\vec{m})\rightarrow\mathbf{D}(\vec{s},\vec{p})$ \\
   7& Darii & some $s$ are $m$, all $m$ are $p$ & some $s$ are $p$  & $\neg\mathbf{D}(\vec{s},\vec{m})\wedge\mathbf{P}(\vec{m},\vec{p})\rightarrow\neg\mathbf{D}(\vec{s},\vec{p})$      \\
   8&Datisi& some $m$ are $s$, all $m$ are $p$ & some $s$ are $p$  & $\neg\mathbf{D}(\vec{m},\vec{s})\wedge\mathbf{P}(\vec{m},\vec{p})\rightarrow\neg\mathbf{D}(\vec{s},\vec{p})$      \\
   9&  Darapti & all $m$ are $s$, all $m$ are $p$ & some $s$ are $p$  & $\mathbf{P}(\vec{m},\vec{s})\wedge\mathbf{P}(\vec{m},\vec{p})\rightarrow\neg\mathbf{D}(\vec{s},\vec{p})$     \\
   10& Disamis & all $m$ are $s$, some $m$ are $p$ & some $s$ are $p$  & $\mathbf{P}(\vec{m},\vec{s})\wedge\neg\mathbf{D}(\vec{m},\vec{p})\rightarrow\neg\mathbf{D}(\vec{s},\vec{p})$      \\
  11& Dimatis & all $m$ are $s$, some $p$ are $m$ & some $s$ are $p$  & $\mathbf{P}(\vec{m},\vec{s})\wedge\neg\mathbf{D}(\vec{p},\vec{m})\rightarrow\neg\mathbf{D}(\vec{s},\vec{p})$      \\
   12& Baroco& some $s$ are not $m$, all $p$ is $m$ & some $s$ are not $p$& $\neg\mathbf{P}(\vec{s},\vec{m})\wedge\mathbf{P}(\vec{p},\vec{m})\rightarrow\neg\mathbf{P}(\vec{s},\vec{p})$     \\
    13& Cesaro& all $s$ are $m$, no $p$ is $m$ & some $s$ are not $p$  & $\mathbf{P}(\vec{s},\vec{m})\wedge\mathbf{D}(\vec{p},\vec{m})\rightarrow\neg\mathbf{P}(\vec{s},\vec{p})$     \\
   14& Celaront & all $s$ are $m$, no $m$ is $p$ & some $s$ are not $p$  & $\mathbf{P}(\vec{s},\vec{m})\wedge\mathbf{D}(\vec{m},\vec{p})\rightarrow\neg\mathbf{P}(\vec{s},\vec{p})$     \\
   15&Camestros& no $s$ is $m$, all $p$ are $m$ & some $s$ are not $p$& $\mathbf{D}(\vec{s},\vec{m})\wedge\mathbf{P}(\vec{p},\vec{m})\rightarrow\neg\mathbf{P}(\vec{s},\vec{p})$     \\
   16& Calemos& no $m$ is $s$, all $p$ are $m$ & some $s$ are not $p$& $\mathbf{D}(\vec{m},\vec{s})\wedge\mathbf{P}(\vec{p},\vec{m})\rightarrow\neg\mathbf{P}(\vec{s},\vec{p})$     \\
   17& Bocardo & all $m$ are $s$, some $m$ are not $p$ & some $s$ are not $p$  & $\mathbf{P}(\vec{m},\vec{s})\wedge\neg\mathbf{P}(\vec{m},\vec{p})\rightarrow\neg\mathbf{P}(\vec{s},\vec{p})$     \\
   18& Bamalip & all $m$ are $s$, all $p$ are $m$ & some $s$ are $p$  & $\mathbf{P}(\vec{m},\vec{s})\wedge\mathbf{P}(\vec{p},\vec{m})\rightarrow\neg\mathbf{D}(\vec{s},\vec{p})$      \\
   19&Ferio& some $s$ are $m$, no $m$ is $p$ & some $s$ are not $p$& $\neg\mathbf{D}(\vec{s},\vec{m})\wedge\mathbf{D}(\vec{m},\vec{p})\rightarrow\neg\mathbf{P}(\vec{s},\vec{p})$     \\
   20&Festino& some $s$ are $m$, no $p$ is $m$ & some $s$ are not $p$& $\neg\mathbf{D}(\vec{s},\vec{m})\wedge\mathbf{D}(\vec{p},\vec{m})\rightarrow\neg\mathbf{P}(\vec{s},\vec{p})$     \\
   21& Ferison& some $m$ are $s$, no $m$ is $p$ & some $s$ are not $p$& $\neg\mathbf{D}(\vec{m},\vec{s})\wedge\mathbf{D}(\vec{m},\vec{p})\rightarrow\neg\mathbf{P}(\vec{s},\vec{p})$     \\
  22&Fresison& some $m$ are $s$, no $p$ is $m$ & some $s$ are not $p$& $\neg\mathbf{D}(\vec{m},\vec{s})\wedge\mathbf{D}(\vec{p},\vec{m})\rightarrow\neg\mathbf{P}(\vec{s},\vec{p})$     \\

   23&Felapton& all $m$ are $s$, no $m$ is $p$ & some $s$ are not $p$& $\mathbf{P}(\vec{m},\vec{s})\wedge\mathbf{D}(\vec{m},\vec{p})\rightarrow\neg\mathbf{P}(\vec{s},\vec{p})$     \\
   24&Fesapo& all $m$ are $s$, no $p$ is $m$ & some $s$ are not $p$& $\mathbf{P}(\vec{m},\vec{s})\wedge\mathbf{D}(\vec{p},\vec{m})\rightarrow\neg\mathbf{P}(\vec{s},\vec{p})$     \\
    \bottomrule
  \end{tabular}
  }
\end{table}

\subsection{Representing family relations}

\begin{table}
  \caption{Representing basic family relations  in ENN}
  \label{familyList}
  \centering
\hspace*{-1.2em}
  \scalebox{0.9}{
  \begin{tabular}{ll|ll|ll}
    \toprule
    relation  &definition    & relation  &definition& relation  &definition\\
    \midrule
  $\mathbf{child}(\vec{e};\vec{p})$   &    $\lceil\vec{e}\rceil\mathbf{P}(\vec{e}, \vec{p})$ & $\mathbf{son}(\vec{e};\vec{p})$   &    $\lceil\vec{e}_\spadesuit\rceil\mathbf{P}(\vec{e}, \vec{p})$  & $\mathbf{daughter}(\vec{e};\vec{p})$   &  $\lceil\vec{e}_\heartsuit\rceil\mathbf{P}(\vec{e}, \vec{p})$    \\
 $\mathbf{parent}(\vec{p};\vec{e})$   &    $\lfloor\vec{p}\rfloor\mathbf{P}(\vec{e}, \vec{p})$ &$\mathbf{father}(\vec{p};\vec{e})$   &   $\lfloor\vec{p}_\spadesuit\rfloor\mathbf{P}(\vec{e}, \vec{p})$   &
  $\mathbf{mother}(\vec{e}, \vec{p})$ & $\lfloor\vec{p}_\heartsuit\rfloor\mathbf{P}(\vec{e}, \vec{p})$      \\
   $\mathbf{spouse}(\vec{h};\vec{w})$   &  $\lfloor\vec{h}\rfloor\mathbf{O}(\vec{h}, \vec{w})$ & $\mathbf{husband}(\vec{h};\vec{w})$   &  $\lfloor\vec{h}_\spadesuit\rfloor\mathbf{O}(\vec{h}, \vec{w})$ &   
$\mathbf{wife}(\vec{w};\vec{h})$&  $\lfloor\vec{w}_\heartsuit\rfloor\mathbf{O}(\vec{h}, \vec{w})$     \\
    \bottomrule
\end{tabular}
}
\end{table}
\begin{table}
  \caption{Representing compound family relations in ENN }
  \label{familyList2}
  \centering
\hspace*{-1.2em}
  \scalebox{0.8}{
  \begin{tabular}{llll}
    \toprule
    formula  & definition & formula  & definition\\
    \midrule
     $\mathbf{sibling}(\vec{p};\vec{e})$   &  $\exists \vec{s} \ [\mathbf{child}(\vec{p};\vec{s})\land\mathbf{child}(\vec{e};\vec{s})]$ &  $\mathbf{grandparent}(\vec{p};\vec{e})$   &   $\exists \vec{m} [\mathbf{parent}(\vec{p};\vec{m})\land \mathbf{parent}(\vec{m};\vec{e})]$  \\
   $\mathbf{niece}(\vec{p};\vec{e})$  & $\exists\vec{s} \ [\mathbf{sibling}(\vec{s};\vec{e})\land\mathbf{daughter}(\vec{p};\vec{s})]$   &$\mathbf{grandson}(\vec{p};\vec{e})$  & $\exists\vec{c} \ [\mathbf{child}(\vec{c};\vec{e})\land\mathbf{son}(\vec{p};\vec{c})]$    \\
  $\mathbf{sonInLaw}(\vec{p};\vec{e})$  & $\exists \vec{d} \   [\mathbf{daughter}(\vec{d};\vec{e})\land\mathbf{husband}(\vec{p};\vec{d})]$  
& $\mathbf{sisterInLaw}(\vec{p};\vec{e})$  & $\exists \vec{b}\  [\mathbf{brother}(\vec{b};\vec{e})\land\mathbf{wife}(\vec{p};\vec{b})]$   
   \\
$\mathbf{aunt}(\vec{p};\vec{e})$ &\multicolumn{3}{l}{$\exists \vec{m} [\mathbf{parent}(\vec{m};\vec{e})\land\mathbf{sister}(\vec{p};\vec{m})]\lor \exists m[\mathbf{parent}(\vec{m};\vec{e})\land\mathbf{sisterInLaw}(\vec{p};\vec{m})]$ }\\
$\mathbf{cousin}(\vec{p};\vec{e})$  & \multicolumn{3}{l}{$\exists\vec{a} \ [\mathbf{aunt}(\vec{a};\vec{e})\land\mathbf{child}(\vec{p};\vec{a})]\lor \exists u [\mathbf{uncle}(\vec{u};\vec{e})\land\mathbf{child}(\vec{p};\vec{u})]$}\\  
    \bottomrule
\end{tabular}
}
\end{table}



Representing and reasoning with family relations is one of the best examples to illustrate the power of neural networks \cite{hinton81}. It is also an example to show spatial thinking (in terms of diagrammatic representation and reasoning) can be applied for non-spatial thinking \cite{tversky19}. We use a ball to represent a family member. The central vector of a ball encodes its latent feature (including gender information); topological relations between balls structure family relations. The lower limit ball $\vec{a}$ satisfying with $\varphi(\vec{a})$, written as `$\lfloor\vec{a}\rfloor \varphi(\vec{a})$', is understood as $\vec{a}$ is the smallest ball that satisfies with $\varphi$, formally, `$\lfloor\vec{a}\rfloor \varphi(\vec{a})\triangleq \varphi(\vec{a})\land\nexists \vec{u}[\mathbf{P}(\vec{u},\vec{a})\land\neg\mathbf{E}(\vec{u},\vec{a})\land\varphi(\vec{u})]$'. The upper limit ball $\vec{a}$ satisfying $\varphi(\vec{a})$, written as `$\lceil\vec{a}\rceil\varphi(\vec{a})$', is understood as $\vec{a}$ is the largest ball that satisfies $\varphi$, formally, `$\lceil\vec{a}\rceil \varphi(\vec{a})\triangleq \varphi(\vec{a})\land\nexists \vec{u}[\mathbf{P}(\vec{a},\vec{u})\land\neg\mathbf{E}(\vec{u},\vec{a})\land\varphi(\vec{u})]$'. Given person $p$, person $e$ being his/her child can be represented as ball $\vec{e}$ is an upper limit ball inside ball $\vec{p}$, written as $\mathbf{Child}(\vec{e};\vec{p})\triangleq \lceil\vec{e}\rceil\mathbf{P}(\vec{e}, \vec{p})$. 
We use $\heartsuit$ and  $\spadesuit$ to represent female and male, respectively. Person $p$ being the mother of person $e$ is written as `$\mathbf{Mother}(\vec{p};\vec{e})\triangleq \lfloor\vec{p}\rfloor\mathbf{P}(\vec{e}, \vec{p})\land \heartsuit(\vec{p})$' or for short `$\lfloor\vec{p}_\heartsuit\rfloor\mathbf{P}(\vec{e}, \vec{p})$'. We introduce an ethnic axiom that siblings should not be married and become spouse.
ENN is able to precisely represent all family relations in English. Table~\ref{familyList}-\ref{familyList2} lists a number of representative family relations, other relations are defined in the similar manner.
\begin{named}[Ethnic Axiom.]
    $\forall \vec{p},\vec{e}\ [\mathbf{sibling}(\vec{p};\vec{e})\rightarrow \mathbf{D}(\vec{p},\vec{e})]$
\end{named}

\section{Experiments}

For all datasets, we set the dimension $N$ of balls as 2 or 3. The ideal spatial values $N_D$, $N_O$, $N_P$ and $M$ are set as 3, 3, 3 and 72, respectively. The maximum number of iterations $maxInter$ is 1000. We leverage the stochastic gradient descent~\cite{bottou2010large} to optimize the spatial relations between balls according to Algorithm~\ref{algo:uenn0}, and the learning rate is chosen as 0.005. We implemented ENN in PyTorch. All experiments were conducted on a personal workstation with an Intel(R) Xeon(R) E5-2640 2.40GHz CPU, and 256 GB memory.

\subsection{Learning syllogism}

\begin{figure} 
\centering
\includegraphics[width=0.7\textwidth]{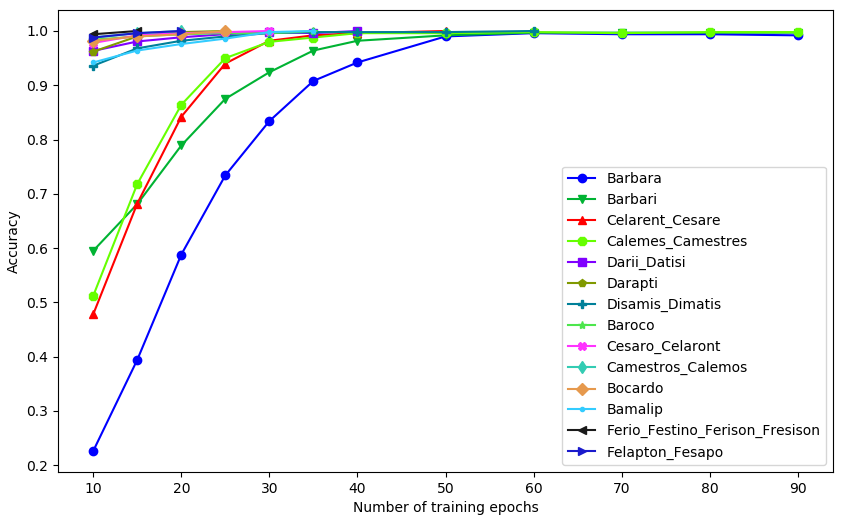} 
\caption{After 60 epochs of training, reasoning accuracies of Syllogism reasoning in structures of Barbara, Barbari, Celarent\_Cesare, Calemes\_Camestres reach to 99.6\%, 99.7\%, 99.8\%, 99.8\%, respectively; accuracies of other Syllogism structures reach to 100.0\% 
 }
\label{fig:syllogism}
\end{figure}

We group 24 syllogism structures into 14 groups. Syllogism structures in the same group can be tested by the same dataset. For each group, we created 500 test cases by extracting hypernym relations of WordNet3.0. A test case consists of 2 assertions as premises, 1 true conclusion, and 1 false conclusion, totaling 14,000 assertions for training, and 7,000 true testing assertions and 7,000 false testing assertions.
As shown in Figure~\ref{fig:syllogism},
ENN achieves the superior accuracy in reasoning with a variety of syllogism structures, and demonstrates great potential, in contrast to traditional neural networks, in reasoning with complex knowledge.

\subsection{Learning Family Relations}

We extracted all Triples of basic family relations (spouse, child, and parent) from DBpedia for training, and created complex family relations without gender information according to Table~\ref{familyList}-\ref{familyList2} for testing. 
We group training Triples into family groups. Two persons are in the same family group, if there are a chain of basic family relations between them. Family groups are sorted by the number of family members. We ignore sorts under which there are less than 5 family groups. The statistics of the dataset, after cleaning, is listed in Table~\ref{fam_data}.  

\begin{table}\caption{Datasets extracted from DBpedia for reasoning with family relations (\#Member: number of family members; \#Family: number of family groups having \#Member members; \#Triple: number of basic relation triples; \#True\_A: number of true assertions; \#False\_A: number of false assertions)}
\label{fam_data}
\centering
\hspace*{-1.2em}
\scalebox{0.8}{
\setlength{\tabcolsep}{5.0pt} 
\begin{tabular}{l|ccccccccccccccc}
\toprule
\#Member & 3 & 4 & 5 & 6 & 7 & 8 & 9 & 10 & 11 & 12 & 13 & 14 & 15 & 16 & 17\\
\#Family & 1,899 & 1,060 & 591 & 344 & 194 & 121 & 69 & 62 & 42 & 28 & 14 & 18 & 8 & 8 & 6\\
\midrule
\#Triple & 3,803 & 3,193 & 2,402 & 1,746 & 1,183 & 876 & 585 & 573 & 438 & 321 & 178 & 251 & 119 & 125 & 98\\
\midrule
\#True\_A & 2,595 & 2,937 & 2,577 & 2,165 & 1,626 & 1,295 & 942 & 912 & 745 & 505 & 308 & 569 & 255 & 208 & 259\\
\#False\_A & 2,595 & 2,944 & 2,630 & 2,202 & 1,649 & 1,351 & 1,023 & 940 & 772 & 527 & 326 & 603 & 265 & 219 & 265\\
\bottomrule
\end{tabular}
}
\end{table}

\begin{figure} 
\centering
\includegraphics[width=0.8\textwidth]{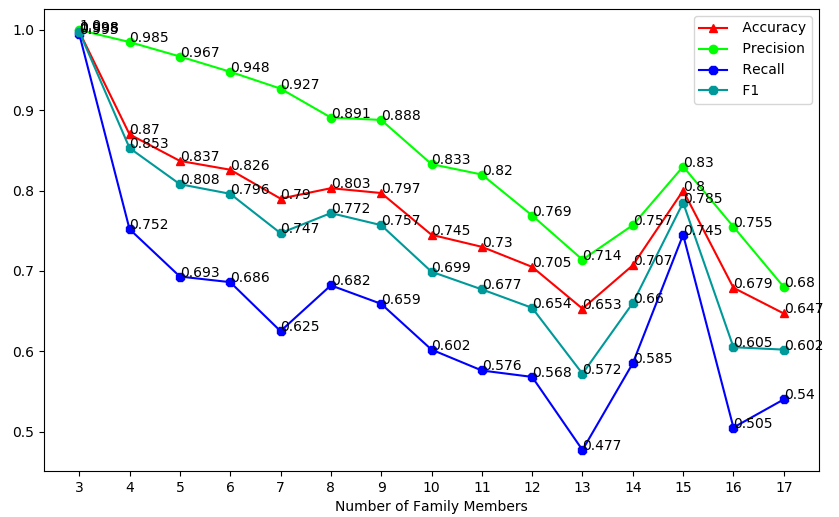} 
\caption{From simple relation with a simple family, ENN can infer cross-family relations
 }
\label{fig:family}
\end{figure}

Experiment results show that for training sets only consists of three persons, the reasoning turns out to be Syllogism, ENN reaches almost 100\% precision and recall. The performance decreases, as the number of family members increases, as illustrated in Figure~\ref{fig:family}. Most errors are resulted from two reasons: (1) the loss of training process fails to reach the global minimum 0 within the maximum number of iteration; (2) there are family members in the dataset that violated the ethnic axiom.

\section{Related work}


Regions, e.g. Venn diagram or Euler diagram, have been used to represent logical reasoning \cite{Venn1880,Stapleton10}, and can be embedded by representation learning. For example, words or entities are embedded as multi-modal Gaussian distribution\cite{w2vGaussian17,transGaussian} or as manifolds \cite{XiaoIjcai16}; nested regions are embedded by Poincar\'e balls to encode tree structures \cite{Nickel17};  Spheres are used to embed concepts to capture subordinate relations among instances and concepts \cite{JuanziEmnlp18}. Intersection or union among high-dimensional boxes are implemented to approximate a subset of logical queries \cite{ren2020}.  Hyperbolic disks are trained to embed directed acyclic graphs (DAGs)\cite{hyperbolicDAG19}. Relations between regions, including distance and orientation, can be logically formalized by taking the connection relation as primitive  and calculated \cite{Laguna22a,whitehead29,Clarke81,Randell92a,Smith94,DongJPL}. The connection relation is valued in the research of cognitive science in the sense that the contact relation \cite{carey09}, or the topological relation \cite{Piaget54}, is the first relation distinguished by human babies. Under uncertain or incomplete situations, 
reasoning will turn out to be similarity judgments \cite{TverskyKahneman74,Tversky77}, which can be approximated by $\cos$ similarity between vectors\footnote{vectors can be understood as regions of the smallest size} \cite{hinton81}.

 \section{Conclusions and Outlooks}

One major challenge for neural reasoning is to reach
the symbolic level of analysis \cite{bechtel02}. Recent studies suggest that pre-trained 
neural language models have a long way to go to adequately learn
human-like factual knowledge \cite{HS20ACL}. 
In this paper, we loose the tie of neural model to vector representations, and propose a novel neural architecture ENN that takes high-dimensional balls as input. 
We show that topological relations among balls is able to spatialize semantics of symbolic logic. ENNs can precisely represent human-like factual knowledge, such as all 24 different structures of Syllogism, and complex family relations. Our experiments show that the novel global optimazation algorithm pushes the reasoning ability of ENN to the level of symbolic syllogism. In ENN, the central vector of a ball is able to inherit the representation power of traditional neural-networks. Jointly training ENN with unstructured and structured data is our on-going research.

\appendix
\section{Parameter setting}

For parameter setting, we choose the dimension $N$ of balls among $\{2, 3, 5, 10\}$ (the results for Syllogism are slightly better when $N=5$). In order to better visualize the ball representations and observe their spatial relation, we set $N=2$ in all datasets.
The ideal spatial values $N_D$, $N_O$, $N_P$ are choosen among $\{2, 3, 4, 5\}$. 
And the ideal spatial values $M$ is choosen among $\{6, 12, 36, 72\}$, which means each time we rotate the ball by $\{60^\circ, 30^\circ, 10^\circ, 5^\circ\}$, respectively.
The maximum number of iterations $maxInter$ is 1000.
The learning rate $\lambda$ is chosen among $\{0.05, 0.01, 0.005, 0.001\}$.
The optimal configuration of ENN for these two datasets is: $N_D=3, N_O=3, N_P=3, M=72, \lambda=0.005$. 
 
\section{Experiments with Syllogism}

\subsection{Datasets}

Assertions in syllogisms take four forms as follows: (1) {\em all a are b}, (2) {\em some a are b}, (3) {\em no a are b}, and (4) {\em some a are not b}. We use four Triple forms: (i) \verb+all a b+, (ii) \verb+some a b+, (iii) \verb+no a b+, and (iv) \verb+some-not a b+, respectively. 
 
 A test case consists of 2 assertions as premises, 1 true conclusion, and 1 false conclusion. We use `,' to separate two premise assertions, `:' to separate premises from the true conclusion, and `;' to separate the true conclusion from the false conclusion. For example,
\begin{verbatim}
	all leader.n.01 person.n.01, 
	all person.n.01 entity.n.01: 
	all leader.n.01 entity.n.01; 
	some-not leader.n.01 entity.n.01
\end{verbatim}

Euler Neural-Network shall learn an Euler diagram from the two premise assertions, and decide whether  truth-values of the two assertions. 
Experiment datasets are extracted from the hypernym relations  of WordNet3.0. We group 24 syllogism structures into 14 groups.  For each group, we created 500 test cases, save them in a file, one line for one test case, totaling 14,000 assertions for training, and 7,000 true testing assertions and 7,000 false testing assertions. 


\subsection{Experiment results}

\subsubsection{Reasoning with sample syllogism structures}

We present some learning examples of ENN. 

\paragraph{Barbara}
Type Barbara takes the form that {\em if all $s$ are $m$, all $m$ are $p$, then all $s$ are $p.$}  Test cases are in the file \verb+Modus_Barbara.txt+, each line has the form \verb+all <x> <y>, all <y> <z>: all <x> <z>; some-not <x> <z>+

\begin{figure}
\begin{subfigure}[c]{0.5\textwidth}
\centering
\includegraphics[width=1.0\textwidth]{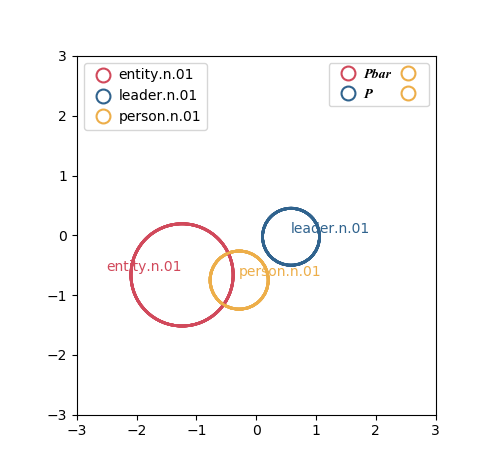}
\subcaption{Random initialization}
\end{subfigure}
\begin{subfigure}[c]{0.5\textwidth}
\centering
\includegraphics[width=1.0\textwidth]{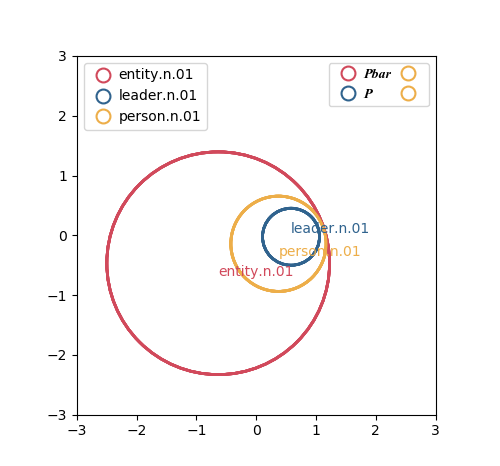}
\subcaption{Learned by ENN}
\end{subfigure}
\caption{Learning Type `Barbara' Syllogism  by ENN}
\label{babara}
\end{figure}

For the test case \begin{verbatim}
	all leader.n.01 person.n.01, 
	all person.n.01 entity.n.01: 
	all leader.n.01 entity.n.01; 
	some-not leader.n.01 entity.n.01
\end{verbatim} our ENN will learn and generate an Euler diagram as illustrated in Figure~\ref{babara}, and correctly concludes: (1) it is true that \verb+all leader.n.01 entity.n.01+; (2) it is false that \verb+some-not leader.n.01 entity.n.01+.

\paragraph{Barbari}

\begin{figure}[h]
\begin{subfigure}[c]{0.5\textwidth}
\centering
\includegraphics[width=1.0\textwidth]{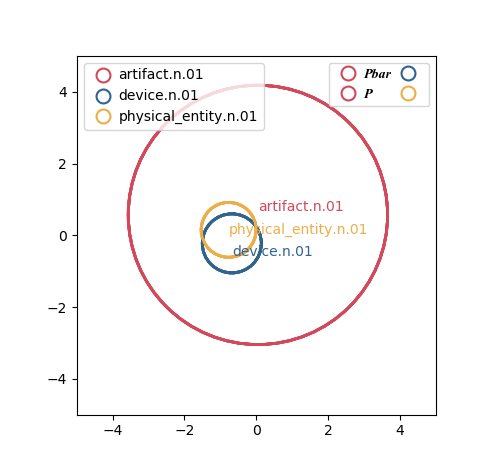}
\subcaption{Random initialization}
\end{subfigure}
\begin{subfigure}[c]{0.5\textwidth}
\centering
\includegraphics[width=1.0\textwidth]{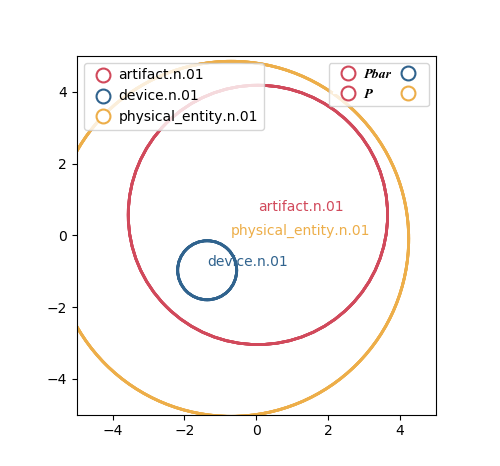}
\subcaption{Learned by ENN}
\end{subfigure}
\caption{Learning Type `Barbari' Syllogism  by ENN}
\label{barbari}
\end{figure}

Type Barbari takes the form that {\em if all $s$ are $m$, all $m$ are $p$, then some $s$ are $p.$} Test cases are in the file \verb+Modus_Barbari.txt+, each line has the form \verb+all <x> <y>, all <y> <z>: some <x> <z>; no <x> <z>+. For the test case 
\begin{verbatim}
	all device.n.01 artifact.n.01, 
	all artifact.n.01 physical_entity.n.01:
	some device.n.01 physical_entity.n.01; 
	no device.n.01 physical_entity.n.01
\end{verbatim}
our ENN will learn and generate an Euler diagram as illustrated in Figure~\ref{barbari}, and correctly concludes: (1) it is true that \verb+some device.n.01 physical_entity.n.01+; (2) it is false that \verb+no device.n.01 physical_entity.n.01+.

\bibliographystyle{plain}
\bibliography{XBib,XBib_NN}

\end{document}